\documentclass[runningheads]{llncs}
\usepackage{graphicx}

\usepackage{tikz}
\usepackage{comment}
\usepackage{amsmath,amssymb} 
\usepackage{color}

\usepackage[accsupp]{axessibility}  

\usepackage{algpseudocode}
\usepackage{algorithm}
\usepackage{graphicx}
\usepackage{subcaption}

\usepackage{listings}
\usepackage{xcolor}
\definecolor{codegreen}{rgb}{0,0.6,0}
\definecolor{codegray}{rgb}{0.5,0.5,0.5}
\definecolor{codepurple}{rgb}{0.58,0,0.82}
\definecolor{backcolour}{rgb}{0.95,0.95,0.92}

\lstdefinestyle{mystyle}{
    backgroundcolor=\color{backcolour},   
    commentstyle=\color{codegreen},
    keywordstyle=\color{magenta},
    numberstyle=\tiny\color{codegray},
    stringstyle=\color{codepurple},
    basicstyle=\ttfamily\scriptsize,
    breakatwhitespace=false,         
    breaklines=true,                 
    captionpos=b,                    
    keepspaces=true,                 
    numbers=left,                    
    numbersep=5pt,                  
    showspaces=false,                
    showstringspaces=false,
    showtabs=false,                  
    tabsize=2
}

\begin{document}
\pagestyle{headings}
\mainmatter

\title{Active Learning for Imbalanced Civil Infrastructure Data} 

\titlerunning{Active Learning for Imbalanced Civil Infrastructure Data}

\author{Thomas Frick\inst{1, 2}
\and
Diego Antognini\inst{1}
\and
Mattia Rigotti\inst{1}
\and
Ioana Giurgiu\inst{1}
\and
Benjamin Grewe\inst{2}
\and
Cristiano Malossi\inst{1}
}

\authorrunning{T. Frick et al.}

\institute{IBM Research, Zurich, Switzerland \\
\email{\{fri,mrg,igi,acm\}@zurich.ibm.com, diego.antognini@ibm.com},\\ 
\and
Institute of Neuroinformatics, UZH and ETH Zurich, Switzerland \\
\email{bgrewe@ethz.ch}}

\maketitle

\begin{abstract}

Aging civil infrastructures are closely monitored by engineers for damage and critical defects. As the manual inspection of such large structures is costly and time-consuming, we are working towards fully automating the visual inspections to support the prioritization of maintenance activities. To that end we combine recent advances in drone technology and deep learning. Unfortunately, annotation costs are incredibly high as our proprietary civil engineering dataset must be annotated by highly trained engineers. Active learning is, therefore, a valuable tool to optimize the trade-off between model performance and annotation costs. Our use-case differs from the classical active learning setting as our dataset suffers from heavy class imbalance and consists of a much larger already labeled data pool than other active learning research. We present a novel method capable of operating in this challenging setting by replacing the traditional active learning acquisition function with an auxiliary binary discriminator. We experimentally show that our novel method outperforms the best-performing traditional active learning method (BALD) by 5\% and 38\% accuracy on CIFAR-10 and our proprietary dataset respectively.

\end{abstract}

\section{Introduction}
Civil infrastructures are constantly being monitored for critical defects, as failure to recognize deficiencies can end in disaster. Unfortunately, detailed inspections are time-consuming, costly, and sometimes dangerous for inspection personnel. We are working towards automating all aspects of visual inspections of civil infrastructures by recent advances in drone technology~\cite{intelligenceDroneTechnologyUses,chanProgressDroneTechnology2018} that enable fast and remote visual inspection of inaccessible structures such as wind turbines, water dams, or bridges and simultaneous advances in instance segmentation using deep learning models~\cite{renGenericDeepLearningBasedApproach2018,davtalabAutomatedInspectionRobotic2022,yinDeepLearningbasedFramework2020} that facilitate accurate detection of surface defects on high resolution images. We are combining these novel technologies by utilizing drones equipped with high-resolution cameras to capture image material of the structures and deep neural networks to detect defects. Our goal is to decrease the duration of inspections, improve condition assessment frequency, and ultimately minimize harm to human health.

It is relatively simple to collect large amounts of data of civil infrastructures for our proprietary dataset. However, similar to many other projects applying deep learning, annotating the collected samples is incredibly costly and time-intensive as highly-trained engineers are required for the labeling process. To minimize costs, a popular approach is to annotate a small portion of the dataset first and then use Active Learning (AL) to select informative samples that should be labeled next. The goal is to optimize a trade-off between additional annotation cost (number of annotated samples) and an increase in model performance. AL has been successfully applied in medical imaging~\cite{shiActiveLearningApproach2019,liPathALActiveLearning2022}, astronomy~\cite{richardsACTIVELEARNINGOVERCOME2011}, and surface defect detection~\cite{fengDeepActiveLearning2017}. For its application in industry, we observe two differences to the traditional AL setting:

First, previous research\cite{galDropoutBayesianApproximation2016a,BayesianActiveLearning2022,kirschBatchBALDEfficientDiverse2019a} has focused on starting the AL process with little to no data (100 samples or cold start). In contrast, industry projects often start the AL process with a larger pool of labeled training data: Despite the high annotation costs, a random set of initial data is labeled for a proof of concept. Only then data collection and labeling efforts are scaled up.  In this phase, the efficient selection of samples is invaluable. Unfortunately, traditional AL strategies barely outperform random selection given a large initial dataset, as shown in our experiments.

Secondly, most academic datasets are class balanced, with each class having the same number of samples (e.g., CIFAR-10~\cite{krizhevskyLearningMultipleLayers} 5,000). In contrast, real-world industry datasets usually suffer from a long-tailed class distribution ~\cite{zhangDeepLongTailedLearning2021,liuLargeScaleLongTailedRecognition2019}. Moreover, the minority classes are often the most important ones. This is usually the case for civil structures: dangerous/critical defects rarely appear on drone scans of the structure as they are well maintained. Previous work on AL for imbalanced data ~\cite{aggarwalActiveLearningImbalanced2020,wangImportantSamplingBased2020,kwolekBreastCancerClassification2019} has shown that classical AL methods fail to select samples of the minority classes for heavily imbalanced datasets and therefore fail to improve model performance for the minority classes. These works focus on developing sample selection strategies that choose samples according to an uncertainty-based metric or a diversity-based approach.

We present a novel method that effectively and efficiently selects minority samples from a pool of unlabeled data for large datasets suffering from heavy class imbalance. Contrary to other AL methods that try to find informative samples from all classes, we limit ourselves to selecting samples for a single minority class, investing the total labeling budget for samples that improve model performance for only that minority class. To this end, our method replaces the AL acquisition function with a binary discriminator explicitly trained in a one-vs-all fashion (minority vs. majority classes) to distinguish between unlabeled minority and majority samples. In each cycle, the discriminator selects samples to be labeled next according to the highest prediction scores. We experimentally confirm that classical active learning methods fail to significantly improve model performance for CIFAR-10~\cite{krizhevskyLearningMultipleLayers} and our proprietary civil engineering dataset~\ref{sec:background}. Applying our method to our proprietary civil infrastructure dataset, we show a minority class recall improvement of 32\% and an overall accuracy gain of 14\% compared to the best-performing traditional AL method (BALD~\cite{houlsbyBayesianActiveLearning2011b}).

\section{Related Work}
\begin{figure}[!t]
\centering
\includegraphics[width=0.7\textwidth]{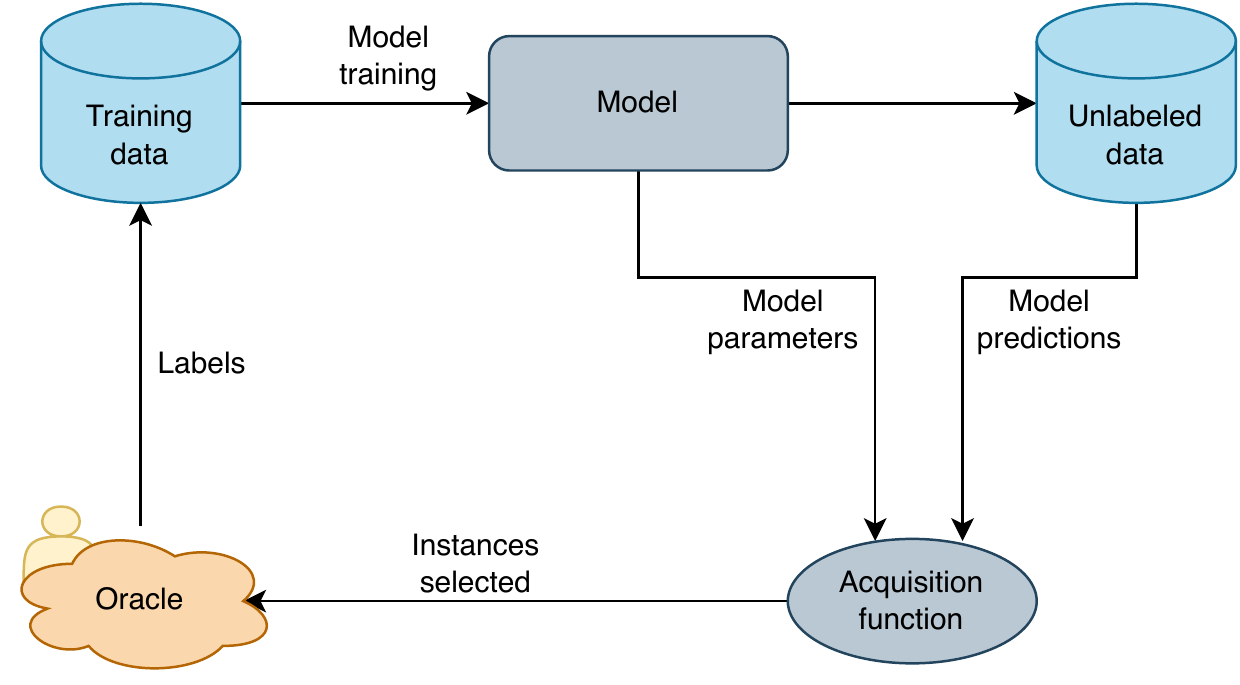}
\caption{Active Learning cycle: a model is trained on a pool of labeled data. Given the model predictions and parameters, an acquisition function selects informative samples from an unlabeled pool of data which are sent to the oracle for labeling.}
\label{fig:active_learning}
\end{figure}

\subsubsection{Active learning} is a well studied problem ~\cite{settlesActiveLearningLiterature2009a,hannekeTheoryDisagreementBasedActive2014,aggarwalActiveLearningImbalanced2020} and especially advantageous for applications with high annotation costs because highly specialized annotators are needed. Active learning works by iterative selecting informative samples according to a query strategy from an unlabeled pool of data. The chosen samples are passed to the oracle (usually a human annotator) to be labeled. The goal is to improve model performance as much as possible while labeling as few samples as necessary. Settles et al.~\cite{settlesActiveLearningLiterature2009a} identifies two different AL scenarios:  In sample-based selective sampling, a stream of samples arrives one after another. Therefore, the algorithm decides whether to label or discard each sample without information about samples arriving in the future. In contrast, in pool-based active learning, the algorithm has access to the entire pool of unlabeled samples and needs to select samples to be labeled next. Our work is set in the second scenario.

\subsubsection{Active Learning query strategies} can be divided into geometric-based and uncertainty-based methods. Geometric-based approaches make use of the underlying feature space to select informative samples. Feature embeddings are necessary to have a meaningful feature space for an application to high-dimensional input data such as images. For our application, we focus on uncertainty-based methods as we assume that there is no pre-trained model available that can be used as an embedding. There is a wide variety of work on uncertainty estimation of deep neural networks \cite{gawlikowskiSurveyUncertaintyDeep2022a}. Two popular methods that estimate uncertainty by generating multiple predictions on the same input are ensembling multiple models~\cite{lakshminarayananSimpleScalablePredictive2017a} or using Dropout as a bayesian approximation~\cite{galDropoutBayesianApproximation2016a}. As the first option is highly compute-intensive, we make use of the second one for this work.

\subsubsection{Uncertainty-based acquisition functions} judge a samples informativeness based on model uncertainty. The Entropy acquisition function~\cite{shannonMathematicalTheoryCommunication} chooses samples that maximise the predictive entropy, BALD~\cite{houlsbyBayesianActiveLearning2011b} selects data points that are expected to maximise the information gained about the model parameters, Variation Ratios~\cite{freemanElementaryAppliedStatistics1965a} measures lack of confidence. BatchBALD~\cite{kirschBatchBALDEfficientDiverse2019a} improves on BALD by selecting sets of samples that are jointly informative instead of choosing data points that are informative individually. As BatchBALD is impractically compute-intensive for large batches and BALD has been shown to outperform other older methods, we focus only on Entropy and BALD for our experiments.

\subsubsection{Imbalanced datasets} have been widely studied (see Kaur et al.~\cite{kaurSystematicReviewImbalanced2020} for an extensive overview). Prior work frequently utilizes resampling techniques (e.g., oversampling and undersampling, or a combination of both) to balance the training data. While Hernandez et al.~\cite{hernandezEmpiricalStudyOversampling2013} show that simple resampling techniques can significantly improve model performance, Mohammed et al.~\cite{mohammedMachineLearningOversampling2020} conclude that undersampling may discard informative majority class samples and therefore decrease majority class performance. There are synthetic sampling methods (SMOTE~\cite{chawlaSMOTESyntheticMinority2002}, ADSYN~\cite{heADASYNAdaptiveSynthetic2008}) that generate new samples by interpolating in the underlying feature space. However, the underlying input feature space is too complex for image data to apply these methods successfully. Weighting samples according to inverse class frequency has also been successfully applied to prioritize underrepresented classes in training. For our work, we exclusively use oversampling as combining duplication of samples with image augmentation techniques results in a much more diverse set of minority samples than weighting the classes. One disadvantage of this method is the resulting much larger training pool and the longer training times compared to the class-weighted approach.

\section{Background}
\label{sec:background}
As mentioned above, we work towards automating visual inspections of civil infrastructures. We have developed an automated inspection pipeline consisting of 4 consecutive stages: First, drones capture grids of high-resolution images, which are later stitched into full image scenes. Next, state-of-the-art deep learning methods analyze the image scenes and highlight defects such as cracks or rust. Finally, the size of detected defects is measured with a precision of 0.1mm. The pipeline's output supports civil engineers in prioritizing further in-person inspections and maintenance activities.

\subsection{Image Data}
\begin{figure}[!t]
\centering
\includegraphics[width=0.9\columnwidth]{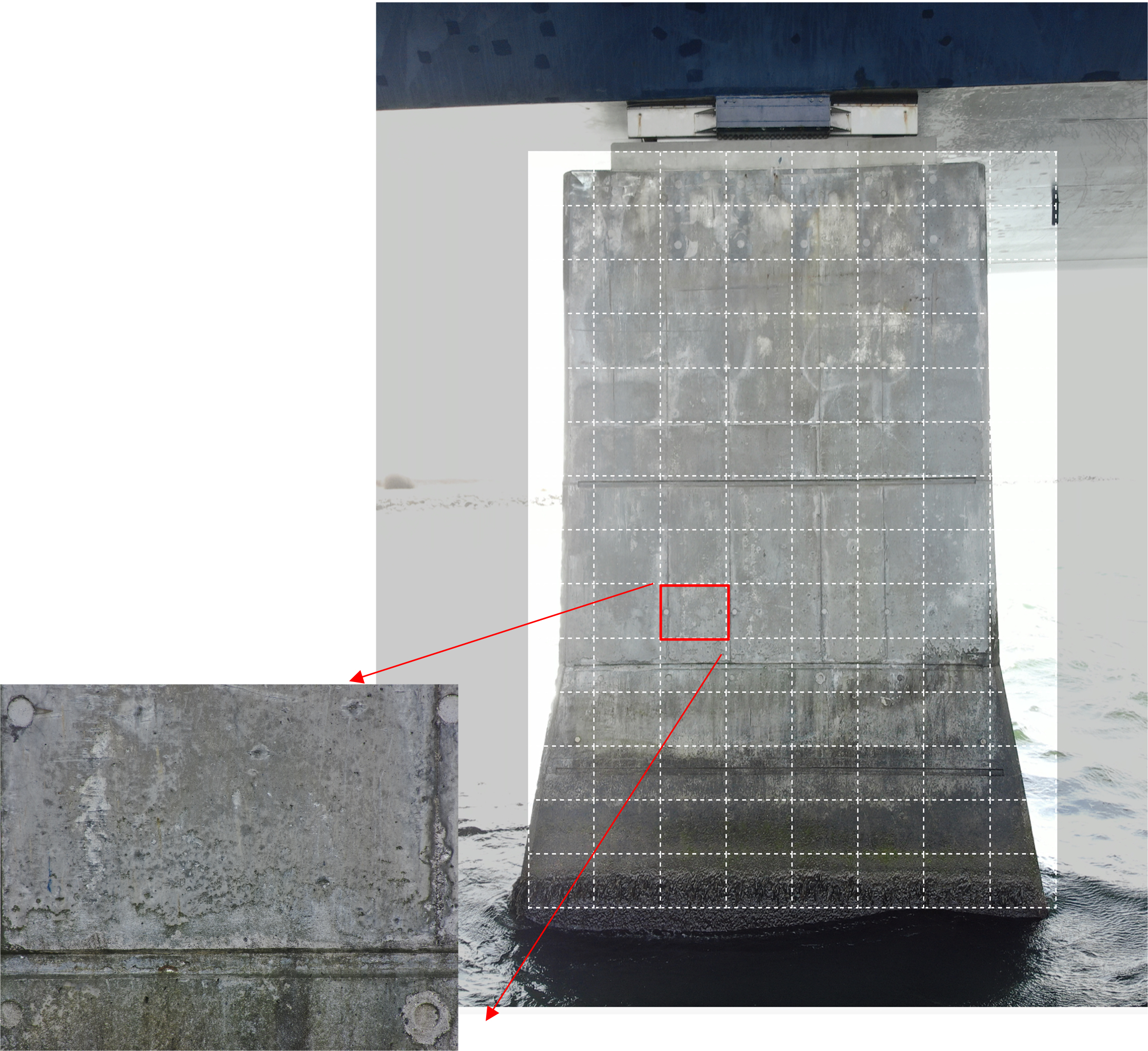}
\caption{DJI Matrix 300 Hi-Res Grid: example of a $7\time14$-grid of 98 high-resolution images. Each individual image is $5184\times3888$ pixel in size resulting in a overall data size for a single pillar face of almost 1GB.}
\label{fig:hi_res_grid}
\end{figure}

We have collected a civil infrastructure dataset consisting of high-resolution images of concrete bridge pillars. The data was collected in 2021 by certified drone pilots using DJI Matrix 300~\cite{Matrice300RTK} drones equipped with a Zenmuse H20 ~\cite{ZenmuseH20Series} lens attachment. Scenes were captured using DJI's Hi-Res Grid Photo mode as shown in Fig. \ref{fig:hi_res_grid}. Drone pilots manually mark the area of the civil structure under inspection on an overview image. While hovering in place, the drone captures a grid of overlapping images by gimbaling the zoom lens. In total, 22 bridge pillars were scanned from each side, resulting in a dataset of over 22'000 raw images ($5184\times3888$ pixels).

\subsection{Instance Segmentation Annotations}
In light of our focus on defect detection, we have invested considerable effort in creating a high-quality instance segmentation dataset where each defect is categorized and localized using mask annotations. In collaboration with our civil engineering experts, we developed extensive annotation guidelines focusing on the following six defect categories: rust, spalling, cracks, cracks with precipitation, net-cracks, and algae (see Fig. \ref{fig:patch_base_algo} for examples). Over six months, a team of annotators labeled 2500 images resulting in around 14'000 defect annotations. Given the well-maintained condition of the inspected structure, critical defects such as rust or cracks with precipitation appear infrequently. Consequently, our dataset suffers from a long-tailed class distribution with a maximum imbalance ratio of 1 to 130 (cracks with precipitation vs. cracks).

\section{Method}
This section first describes our novel method which replace the acquisition function with a binary discriminatory trained on the labeled pool to select minority class samples from the unlabeled pool. Secondly, it describes the dataset conversion algorithm which we use to convert our proprietary instance segmentation dataset to a classification dataset for the experiments.

\subsection{Active Learning for heavily imbalanced data}
The goal of the classical AL setting is to improve the model performance by labeling additional informative samples. A model is trained on the initial pool of labeled data. Given the model's predictions and parameters, an acquisition function determines which samples from the unlabeled pool are sent to the oracle for labeling. The newly annotated samples are then added to the labeled pool, after which the cycle begins again by retraining the model. Traditional AL methods ~\cite{galDropoutBayesianApproximation2016a,houlsbyBayesianActiveLearning2011b,BayesianActiveLearning2022,kirschBatchBALDEfficientDiverse2019a} strive to select samples for which the model performance for all classes improves most.

In contrast, we focus on improving model performance for a single pre-selected minority class. Given the high class imbalance of the initial AL dataset, we hypothesize that we do not need to find the most diverse or informative set of samples but that selecting and labeling any samples of the minority class will improve model performance. Therefore, our method replaces the traditional uncertainty-based AL acquisition function with an auxiliary binary classifier acting as a discriminator between the minority and majority classes. The discriminator is trained on the labeled AL pool in each cycle. Its predictions on the unlabeled pool are used to select samples for labeling. Selection is based on prediction scores, choosing the top-$K$ samples.

For the training of the binary discriminator, binary labels are computed from the original multi-class training set in a one-vs-all fashion. As a result, the binary classification dataset is even more imbalanced than the original multi-class dataset. We improve discriminator performance by combating the bias of the class imbalance by applying the following modifications to the training procedure:
\begin{itemize}
    \item  oversample the positive class (the original minority class) until we reach balance with respect to the negative class (all majority classes);
    \item apply standard image augmentation techniques (flip, shift, scale, rotate, brightness, and contrast - see Albumentations~\cite{buslaevAlbumentationsFastFlexible2020}). As we apply the augmentations to the large number of minority samples generated from the oversampling in the first step, we end up with a highly diverse set of minority class samples; \item apply batch augmentations (MixUp~\cite{zhangMixupEmpiricalRisk2018}, CutMix~\cite{yunCutMixRegularizationStrategy2019} - see timm~\cite{wightmanRwightmanPytorchimagemodelsV02022} library) to further diversify the samples in each batch stabilizing the training procedure of the binary discriminator.
\end{itemize}

\subsection{Instance segmentation to classification dataset conversion}
\label{sec:patchification}
As mentioned in section \ref{sec:background}, we are working towards automated detection of surface defects for civil infrastructures. We train instance segmentation models on the original dataset with instance mask annotations to detect the defects. Our annotations must be labeled with extraordinary precision due to the defects being of such small size (e.g., cracks in the order of millimeters). This additional time effort adds to the already high prize per annotation. Therefore, we have been working on a technique to use weakly supervised learning with class-level supervision to generate class activation maps (CAMs~\cite{selvarajuGradCAMVisualExplanations2020}) from which we extract fine segmentation masks. Unfortunately, there are still high annotation costs associated with class-level labels. We are trying to decrease these costs with active learning in this work. Consequently, we use class-level labels for our experiments which we extract from the original instance segmentation dataset using a patch-based dataset conversion algorithm.

The conversion algorithm works by extracting fixed-size patches from the original images. Intuitively, a patch can be assigned to a category if it depicts a piece of the original class; this can be either shape or texture. While the texture is informative enough for some objects to attribute the correct class, others are only correctly classifiable with information about their shape. As extracted patches only offer a small window into the original image, an object's shape information can be easily lost if a too small patch size is chosen.

\begin{figure}[!t]
\includegraphics[width=\columnwidth]{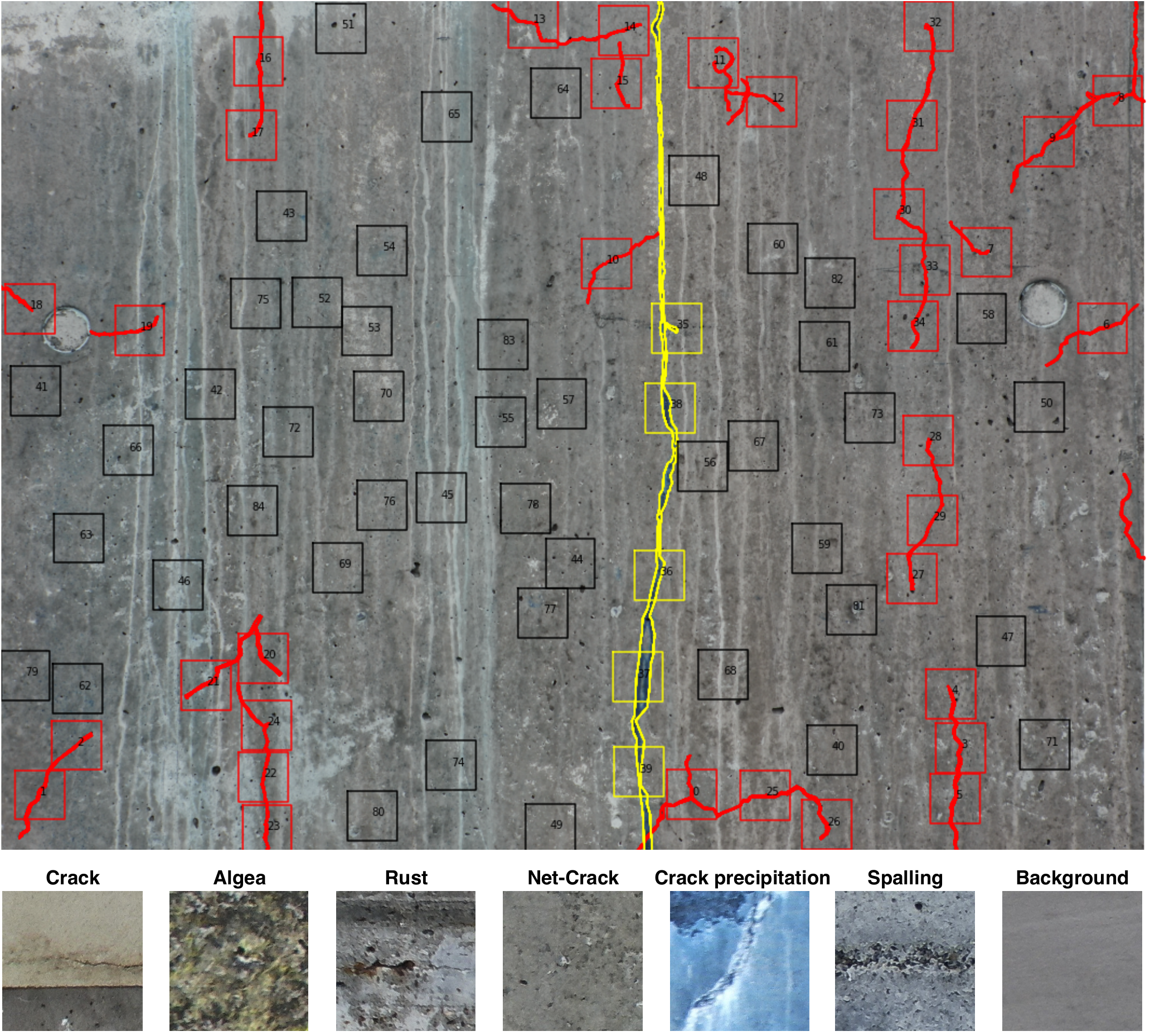}
\caption{Patch-based classification dataset conversion algorithm: we show instance segmentation polygons of the original image colored by class (red crack, yellow crack with precipitation, black background) and sampled patch boarders with the same color scheme. Below, we show one example patch per category of the original dataset plus one example patch for the additional background class.}
\label{fig:patch_base_algo}
\end{figure}

The algorithm processes one image at a time: multiple patches are sampled for each instance and assigned the corresponding category. Additionally, patches are extracted from regions of the original image that do not contain any instances/defects and are assigned to an additional, newly introduced class ``Background''. We randomly sample instance patches such that the center point of the patch lies within the instance annotation and background patches such that the center point lies anywhere within the original image's borders. The algorithm rejects a sampled patch if it violates one of the following criteria: 1) it overlaps with an already chosen valid patch, 2) it intersects with an instance annotation belonging to a different category, 3) the patch breaches the boundaries of the original image. The algorithm extracts patches until the total number of required patches is reached or until the algorithm exceeds a total number of sampling attempts. Figure~\ref{fig:patch_base_algo} shows the sampled patches for an example image as well as one example patch per category.

\begin{figure}[!t]
     \centering
     \begin{subfigure}[b]{0.49\textwidth}
         \centering
         \includegraphics[width=\textwidth]{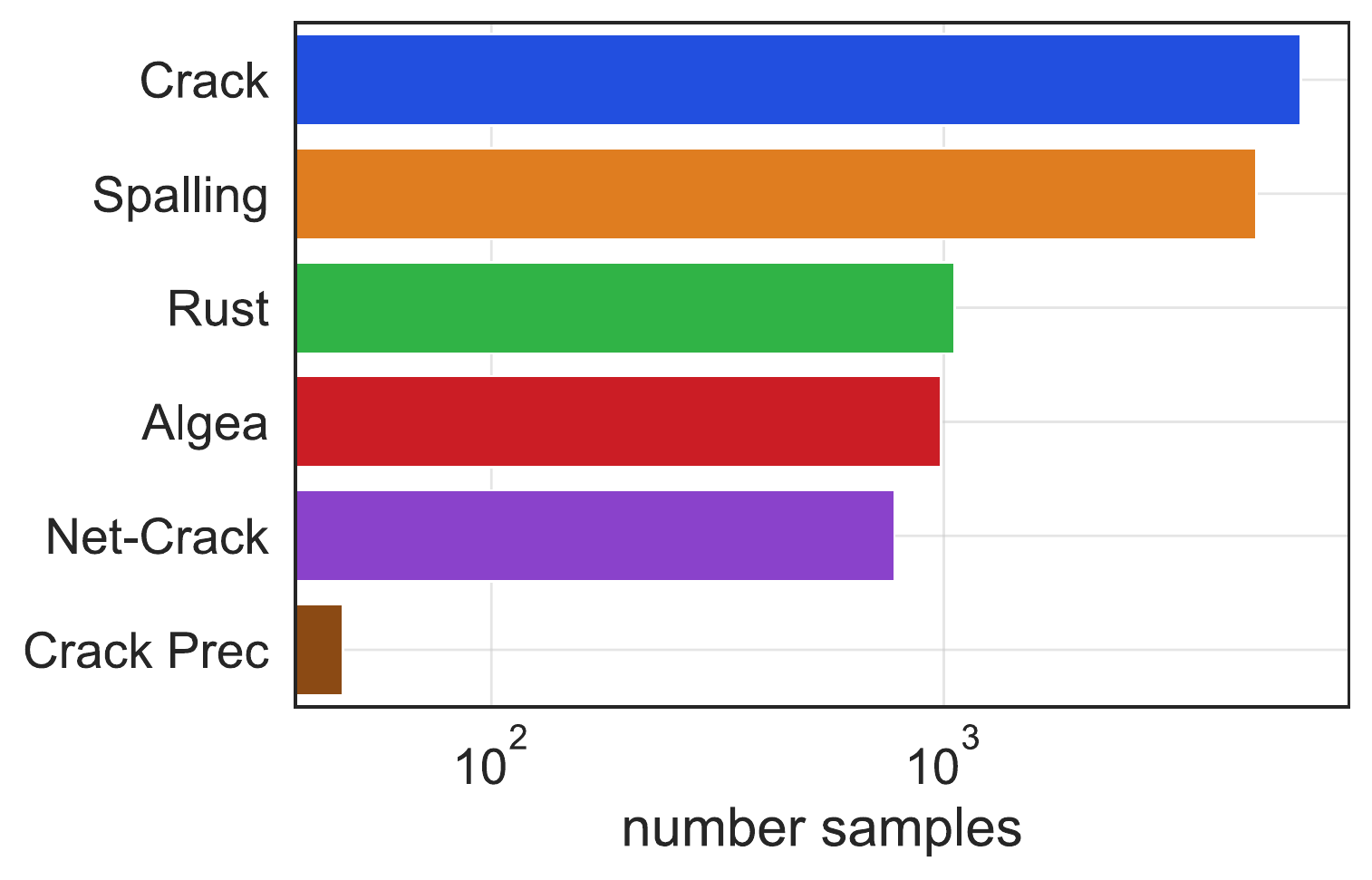}
         \caption{Instance segmentation dataset}
         \label{fig:class_distribution_a}
     \end{subfigure}
     \hfill
     \begin{subfigure}[b]{0.49\textwidth}
         \centering
         \includegraphics[width=\textwidth]{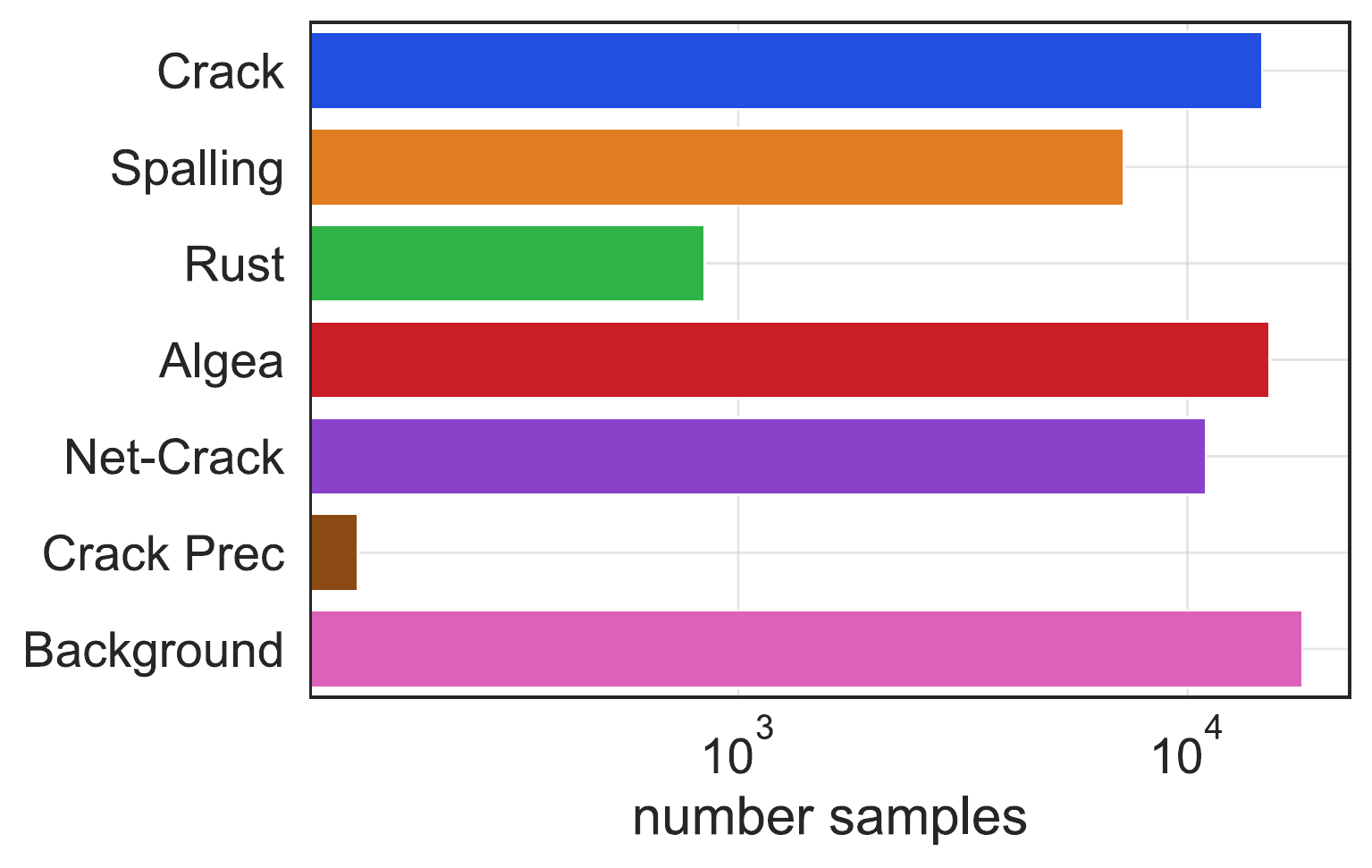}
         \caption{Extracted classification dataset}
         \label{fig:class_distribution_b}
     \end{subfigure}
    \caption{Class distribution of our civil infrastructure datasets: (a) shows statistics for our original instance segmentation dataset while (b) describes statistics for the classification dataset extracted with the path-based algorithm from the instance segmentation dataset (a). The additional background class in Figure (a) is a result of the background patch sampling of the algorithm.}
    \label{fig:class_distribution}
\end{figure}

\section{Evaluation}
In this section we first describe how we convert the image segmentation to a classification dataset with our patch extraction method to prepare our proprietary dataset. Then, we experimentally show how traditional active learning methods fail to improve minority class performance as the initial training data size increases. Finally, we run experiments showcasing model performance for traditional active learning methods, as well as our proposed method on CIFAR-10 and on the civil engineering classification dataset.

\subsection{Civil infrastructure classification dataset}\label{ce_classification_dataset}
We apply our instance segmentation to classification dataset conversion algorithm (see section \ref{sec:patchification}), extracting $160\times160$ pixel patches. Per image, the algorithm attempts to sample $100$ class-patches and $10$ background-patches with a maximum of $100$ sampling attempts. The resulting classification dataset consists of a total of 67'162 samples. Fig. \ref{fig:class_distribution} shows the class distribution for the original instance segmentation dataset and for the extracted classification dataset. Both suffer from heavy class imbalance.

We need a large enough dataset for our experiments to simulate a real-world active learning scenario with a small labeled and a large unlabeled pool. Accordingly, each category included in the final dataset has to contain enough samples such that it is possible to introduce artificial imbalance for each class during our experiments. Therefore, we remove all classes with less than 5'000 samples (crack with precipitation, rust). Additionally, we remove the net-crack class due to the visual similarity of its patches with the crack class. The final dataset consists of four classes: background, algae, crack, and spalling. It is split in a stratified fashion into 70\% training set and 30\% test set.

\subsection{Experiment setup}
\subsubsection{Active learning datasets}
We aim to simulate a real-world active learning scenario with a small pool of labeled data and a large pool of unlabeled data, as well as an oracle that can be queried for labels. Artificial class imbalance is introduced into the labeled and the unlabeled pool to simulate the imbalanced dataset setting dependent on the experiment. For the experiments, the original training set is consequently randomly split into a small labeled set, a large unlabeled set, and an unused set. As we still have access to the labels of the samples of the unlabeled pool, we can simulate human annotation when the active learning algorithm queries labels. Furthermore, once samples have been moved from the unlabeled pool to the labeled pool (simulated labeling process), we can simulate a much larger unlabeled pool by moving the same number of samples per class from the unused pool to the unlabeled pool, restoring the original unlabeled data pool size and class balance. Finally, the test set is used as-is to evaluate the models after each time the AL algorithm queries new samples.

\subsubsection{Model training}
We use the full ResNet18~\cite{heDeepResidualLearning2015,wightmanRwightmanPytorchimagemodelsV02022} as a model backbone for our civil infrastructure dataset, but only use the first ResNet block for CIFAR-10~\cite{krizhevskyLearningMultipleLayers}. All models are trained for 50 epochs to convergence with the AdamW~\cite{loshchilovDecoupledWeightDecay2019} optimizer with a learning rate of $0.0005$ with the sample and batch augmentations highlighted above. For a fair comparison between the traditional AL methods and ours, we train all models with the improved training procedure of oversampling, image and batch augmentations.

\subsubsection{Active learning process}
We apply our method and traditional AL algorithms to each dataset under investigation.
AL algorithms query labels five times for 200 samples per cycle. Each experiment is repeated three times selecting a class from the original dataset as the minority class for the artificially introduced imbalance procedure. All experiments use different randomly initialised model weights and a different random split of the data (labeled and unlabeled pool with artificial imbalance).

\subsection{Results}
\begin{figure}[!t]
    \centering
    \includegraphics[width=0.6\columnwidth]{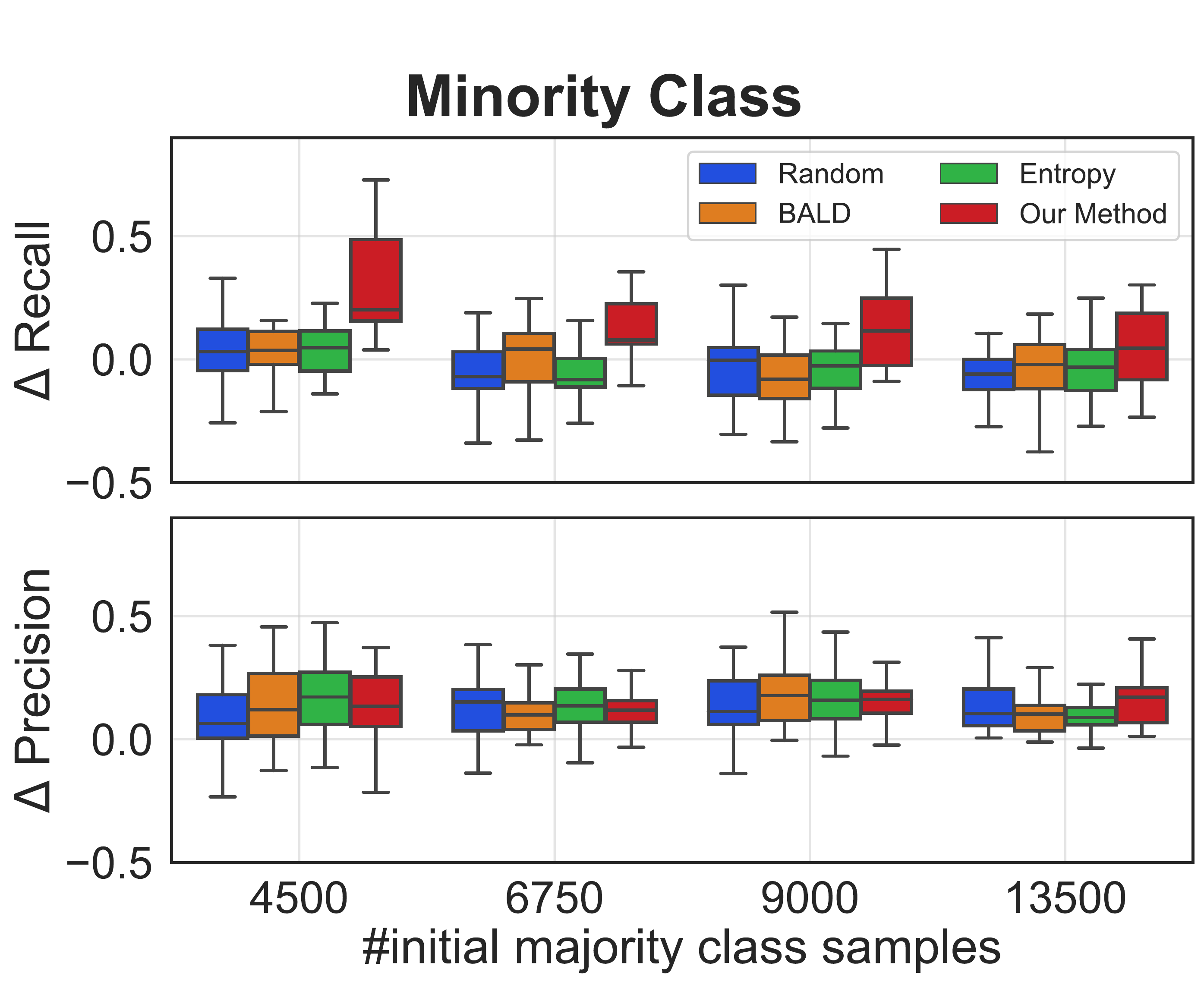}
    \caption{Influence of initial AL training pool size: we report relative minority class performance gains for four dataset sizes with an increasing number of majority samples in the initial AL training pool. Performance differences are evaluated for each method, comparing model performance before the AL procedure with model performance after the procedure queried labels for 1000 samples. While performance gains of traditional AL methods decrease as the training dataset grows, our method retains more of the original performance increase.}
    \label{fig:results_sweep}
\end{figure}

\subsubsection{Influence of initial training pool size}
As a first experiment we evaluate all methods on CIFAR-10~\cite{krizhevskyLearningMultipleLayers} with increasing number of samples in the initial labeled pool. We create artificially imbalanced training datasets with a increasing number of samples for the majority class from 500 to 1500 while keeping the number of minority class samples constant at 50. To minimize the impact of the unlabeled set, it is kept constant with 120 minority samples and 3000 majority samples. Finally, we run the AL cycle for five rounds, selecting and labeling 200 samples per cycle. We measure the change in performance from the initial trained model to the model at the end of the AL procedure. As can be seen in Fig. \ref{fig:results_sweep}, the delta in minority class performance drops off significantly for traditional active learning methods as the initial training dataset size increases. In contrast, our method retains more of the original performance gains as it is less sensitive to the initial data pool size. Specifically, the recall delta remains above zero as initial data size decreases while the classical AL method fall below zero, signaling a decrease in performance from the initial training set to the set with the additional 1000 labeled samples.

\subsubsection{CIFAR-10}
\begin{figure}[!t]
     \centering
     \begin{subfigure}[b]{\textwidth}
         \centering
         \includegraphics[width=\textwidth]{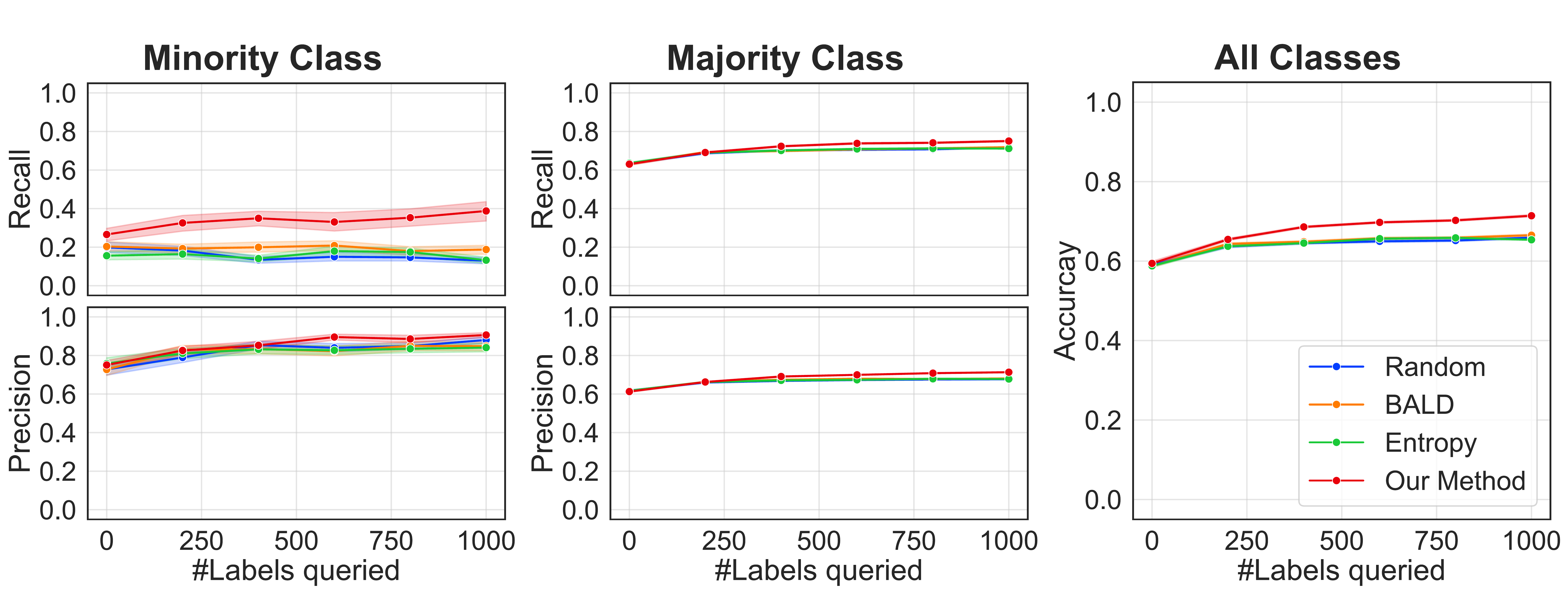}
         \caption{CIFAR-10}
         \label{fig:fig:results_cifar10}
     \end{subfigure}
     \newline
     \begin{subfigure}[b]{\textwidth}
         \centering
         \includegraphics[width=\textwidth]{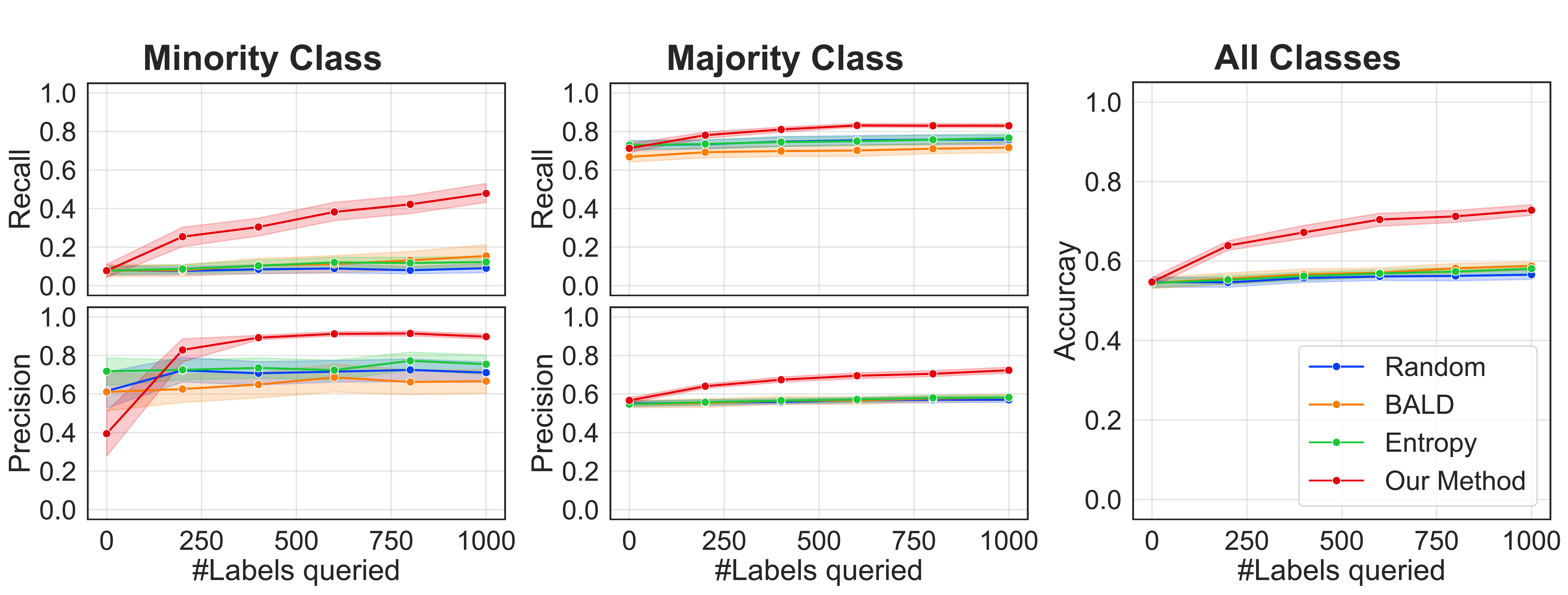}
         \caption{Civil engineering dataset}
         \label{fig:results_ce}
     \end{subfigure}
    \caption{Absolute model performance throughout the AL process: for each cycle, after labeling 200 additional samples, we report precision and recall for the minority class, macro average precision and recall for the majority classes, and overall accuracy for the CIFAR-10 dataset and our proprietary civil infrastructure dataset. Error bands show the standard error of the mean (SEM).}
    \label{fig:results}
\end{figure}

Next, we evaluate absolute model performance on CIFAR-10~\cite{krizhevskyLearningMultipleLayers}, comparing traditional AL methods (BALD~\cite{houlsbyBayesianActiveLearning2011b}, Entropy~\cite{shannonMathematicalTheoryCommunication}) with our novel method on model performance. Additionally, we also include a random acquisition function as a baseline. We create an initial AL dataset that is randomly split into unlabeled and labeled pool for each experiment with an artificial class imbalance of 50 minority samples to 1000 majority samples per class. The unlabeled pool consists of 300 minority samples and 3000 majority samples per class. Results in Fig. \ref{fig:results}a show that traditional AL methods fail to significantly improve precision and recall for the minority class. Additionally, little performance improvement can be seen for the majority classes. We explain this with our experiments' much larger initial training pool compared to other publications. Our initial training set consists of few minority samples but a considerable amount of majority samples. The 200 additionally labeled samples per cycle do not yield much additional information compared to the existing larger training pool. Therefore, neither majority nor minority class performance improves. In contrast, our method focuses only on the minority class, for which only a few samples are in the initial training pool. Therefore, even a few minority samples yield enough information to improve minority class performance considerably. Our method shows a clear performance improvement compared to traditional AL methods. Compared to the random baseline, our method improves on average by 25\% recall of the minority class, while BALD only improves by 5\% and the entropy method only improves by 0.3\%. As a result, the overall accuracy of the model also increases significantly: 5\% average improvement over the random baseline for our method, compared to only 0.6\% for BALD and a 0.5\% decrease for the entropy method.

\subsubsection{Civil engineering dataset}
We evaluate absolute model performance on our proprietary civil engineering dataset (see section \ref{ce_classification_dataset}).  We run experiments on an initial AL dataset with 50 minority and 1000 majority samples per class in the training set. The unlabeled pool consists of 300 minority and 2500 majority samples per class. As with CIFAR-10, the results in Fig. \ref{fig:results_ce} show that traditional AL methods fail to improve precision and recall for the minority class. Meanwhile, our method shows clear improvement in minority class recall and precision as well as overall accuracy. Compared to the results on CIFAR-10, we see a larger performance improvement: compared to the random baseline our method improves by 38\% minority recall, 18\% minority precision, and 16\% overall accuracy. BALD only improves by 6\% minority recall and 2\% overall accuracy while minority precision decreases by 4\%. The entropy method improves only by 3\% minority recall, 4\% minority precision and 1\% overall accuracy. We explain the higher overall accuracy gains compared to CIFAR-10 with the smaller number of classes in our dataset (4 compared to 10 in CIFAR-10). Given the overall accuracy is an average of class-specific performance, minority class improvement has a much larger impact on the majority classes.

\section{Conclusion}
We have presented a novel active learning method that replaces the traditional acquisition function with an auxiliary binary discriminator allowing the selection of minority samples even for initially large and imbalanced datasets. We have experimentally shown that our method outperforms classical AL algorithms on artificially imbalanced versions of CIFAR-10 and our proprietary civil engineering dataset when evaluated on minority class recall, precision, and overall classification accuracy. Consequently, our method facilitates the successful discovery and labeling of rare defects in the yet unlabeled pool of samples for our proprietary civil engineering dataset. Trained on the additional labeled data, our visual inspection defect detection models improve at supporting civil engineers' maintenance prioritization decisions for rare but critical defects.

Due to the large number of models trained per experiment, we are limited to dataset consisting of small images with many classes or large images with few classes. Additionally, our choices of datasets for experimentation were limited as we required many samples per class to simulate a large unlabeled pool. This excluded many popular large datasets as they often consist of many classes with a moderate amount of samples per class.

While we focus on classification datasets in this work, future research could extend our method to an application for the full instance segmentation dataset by creating sliding window patches and aggregating statistics over the full original image. Additionally, it would be interesting to develop our method further to work with multiple minority classes at a time. Finally, research should compare our method with feature space diversity-based AL methods when a pre-trained model (transfer learning or self-supervised learning) is available.

\subsubsection*{Acknowledgement}
This work would not have been possible without Finn Bormlund and Svend Gjerding from Sund\&Bælt. We would like to thank them for their collaboration, specifically for the collection of image data, for their expert annotations, and their tireless help with the annotation guidelines for the civil engineering dataset.

\clearpage
%
%
\bibliographystyle{splncs04}

\end{document}